\documentclass[utf8]{frontiersSCNS} 

\usepackage{url,lineno,microtype,subcaption}
\usepackage[onehalfspacing]{setspace}

\usepackage{nameref}
\usepackage{varioref}
\usepackage{graphicx}
\usepackage{url}
\usepackage{multirow}
\usepackage{multicol}
\usepackage{booktabs}
\usepackage{xcolor}
\usepackage{enumerate}
\usepackage{enumitem}

\def\keyFont{\fontsize{8}{11}\helveticabold }
\def\firstAuthorLast{Kandimalla {et~al.}} 
\def\Authors{Bharath Kandimalla\,$^{1,*}$, Shaurya Rohatgi\,$^{2}$, Jian Wu\,$^{3,*}$ and C. Lee Giles\,$^{1,2}$}
\def\Address{$^{1}$Computer Science and Engineering, Pennsylvania State University, University Park, PA, USA\\
$^{2}$Information Sciences and Technology, Pennsylvania State University, University Park, PA, USA\\
$^{3}$Computer Science, Old Dominion University, Norfolk, VA, USA}
\def\corrAuthor{Bharath Kandimalla , Jian Wu}
\def\corrEmail{bharathkumarkandimalla@gmail.com, jwu@cs.odu.edu}

\begin{document}
\onecolumn
\firstpage{1}
\title[Subject Category Classification]{Large Scale Subject Category Classification of Scholarly Papers with Deep Attentive Neural Networks} 

\author[\firstAuthorLast ]{\Authors} 
\address{} 
\correspondance{} 

\extraAuth{}

\maketitle

\begin{abstract}

Subject categories of scholarly papers generally refer to the knowledge domain(s) to which the papers belong, examples being computer science or physics. Subject category information can be used for building faceted search for digital library search engines. This can significantly assist users in narrowing down their search space of relevant documents. Unfortunately, many academic papers do not have such information as part of their metadata. Existing methods for solving this task usually focus on unsupervised learning that often relies on citation networks. However, a complete list of papers citing the current paper may not be readily available. In particular, new papers that have few or no citations cannot be classified using such methods. Here, we propose a deep attentive neural network (DANN) that classifies scholarly papers using only their abstracts. The network is trained using 9 million abstracts from Web of Science (WoS). We also use the WoS schema that covers 104 subject categories. 
The proposed network consists of two bi-directional recurrent neural networks followed by an attention layer. We compare our model against baselines by varying the architecture and text representation. Our best model achieves micro-${F_1}$ measure of $0.76$ with $F_1$ of individual subject categories ranging from $0.50$--$0.95$. The results showed the importance of retraining word embedding models to maximize the vocabulary overlap and the effectiveness of the attention mechanism. The combination of word vectors with TFIDF outperforms character and sentence level embedding models. We discuss imbalanced samples and overlapping categories and suggest possible strategies for mitigation. We also determine the subject category distribution in CiteSeerX by classifying a random sample of one million academic papers.

\tiny
 \keyFont{ \section{Keywords:}text classification, text mining, scientific papers, digital library, neural networks, CiteSeerX, subject category} 
\end{abstract}

\twocolumn \noindent 

\section{Introduction}
A recent estimate of the total number of English research articles available online was at least 114 million \citep{khabsa2014plosone}. Studies indicate the number of academic papers doubles every 10--15 years \citep{larsen2010}. The continued growth of scholarly papers can make finding relevant research papers challenging. Searches based on only keywords may no longer be the most efficient method \citep{matsuda1999wwwretrieval} to use. This often happens when the same query terms appear in multiple research areas. For example, querying ``neuron'' in Google Scholar returns documents in both computer science and neuroscience. Search results can also belong to diverse domains when the query terms contain acronyms. For example, querying ``IR'' returns documents in Computer Science (meaning ``information retrieval'') and Physics (meaning ``infrared''). Similarly, querying ``NLP'' returns documents in Linguistics (meaning ``neuro-linguistic programming'') and Computer Science (meaning ``natural language processing''). This is because documents in multiple subject categories (SCs) are often mixed together in a digital library search engine and its corresponding SC metadata is usually not available in the existing document metadata, either from the publisher or from automatic extraction methods. 

As such, we believe it can be useful to build a classification system that assigns scholarly papers to SCs. Such an system could significantly impact scientific search and facilitate bibliometric evaluation. It can also help with Science of Science research \citep{fortunato2018science}, a recent area of research that uses scholarly big data to study the choice of scientific problems, scientist career trajectories, research trends, research funding, and other research aspects. 
Also, many have noted that it is difficult to extract SCs using traditional topic models such as Latent Dirichlet Allocation (LDA), since it only extracts words and phrases present in documents\citep{gerlach2018network,agrawal2018wrong}. 
An example is that a paper in computer science is rarely given that label in the keyword(s). In contrast, SC classification is usually based on a universal schema for that a specific domain or for all domains such as that of the Library of Congress. A crowd sourced schema can be found in the DBpedia\footnote{\url{https://en.wikipedia.org/wiki/DBpedia}} of Wikipedia.

In this work, we pose the SC problem as one of multiclass classification in which {\em one} SC is assigned to each paper. In a preliminary study, we investigated feature-based machine learning methods to classify research papers into 6 SCs \citep{wu2018bigdata}.
Here, we extend that study and propose a system that classifies scholarly papers into 104 SCs \emph{using only a paper's abstract}. The core component is a supervised classifier based on recurrent neural networks trained on a large number of labeled documents that are part of the WoS database. In comparison with our preliminary work, our data is more heterogeneous (more than 100 SCs as opposed to 6), imbalanced, and complicated (data labels may overlap). We compare our system against several baselines applying various text representations, machine learning models, and/or neural network architectures. 

Many schemas for scientific classification systems are publisher domain specific. For example, ACM has its own hierarchical classification system\footnote{\url{https://www.acm.org/about-acm/class}}, NLM has medical subject headings\footnote{\url{https://www.ncbi.nlm.nih.gov/mesh}}, and MSC has a subject classification for mathematics\footnote{\url{http://msc2010.org/mediawiki/index.php?title=Main_Page}}. The most comprehensive and systematic classification schemas seem to be from WoS\footnote{\url{https://images.webofknowledge.com/images/help/WOS/hp_subject_category_terms_tasca.html}} and the Library of Congress (LOC)\footnote{\url{https://www.loc.gov/aba/cataloging/classification/}}. The latter was created in 1897 and was driven by practical needs of the LOC rather than any epistemological considerations and is most likely out of date.

To the best of our knowledge, our work is the first example of using a neural network to classify scholarly papers into a comprehensive set of SCs. Other work focused on unsupervised methods and most were developed for specific category domains. In contrast, our classifier was trained on a large number of high quality abstracts from the WoS and can be applied directly to abstracts without any citation information. We also develop a novel representation of scholarly paper abstracts using ranked tokens and their word embedding representations. This significantly reduces the scale of the classic Bag of Word (BoW) model. We also retrained FastText and GloVe word embedding models using WoS abstracts. The subject category classification was then applied to the CiteSeerX collection of documents.  However, it could be applied to any similar collection.

\section{Related Work}
Text classification is a fundamental task in natural language processing. Many complicated tasks use it or include it as a necessary first step, e.g., part-of-speech tagging, e.g., \citep{ratnaparkhi1996maximum}, sentiment analysis, e.g., \citep{vo2015target}, and named entity recognition, e.g., \citep{nadeau2007survey}.
Classification can be performed at many levels: word, phrase, sentence, snippet  (e.g., tweets, reviews), articles  (e.g., news articles), and others. 
The number of classes usually ranges from a few to nearly 100. 
Methodologically, a classification model can be supervised, semi-supervised, and unsupervised. An exhaustive survey is beyond the scope of this paper. Here we briefly review short text classification and highlight work that classifies scientific articles.

Bag of words (BoWs) is one of the most commonly used representations for text classification, an example being keyphrase extraction \citep{caragea2016aaai,he2018keyphrase}. BoW represents text as a set of unordered word-level tokens, without considering syntactical and sequential information. TF-IDF is commonly used as a measure of importance \citep{baezayates1999moden}. Pre-extracted topics (e.g., LDA) have also been used to represent documents before supervised classification \citep{llewellyn2015tpdl}. 

Recently, word embeddings (WE) have been used to build distributed dense vector representations for text. Embedded vectors can be used to measure semantic similarity between words \citep{mikolov2013w2v}. WE has shown improvements in semantic parsing and similarity analysis, e.g., \citet{prasad2018nparscit,bunk2018welda}. Other types of embeddings were later developed for character level embedding \citep{zhang2015nips}, phrase embedding \citep{passos2014lexicon}, and sentence embedding \citep{cer2018universal}. Several WE models have been trained and distributed; examples are word2vec \citep{mikolov2013w2v}, GloVe \citep{pennington2014glove}, FastText \citep{grave2017fasttext}, Universal Sentence Encoder \citep{cer2018universal}, ELMo \citep{peters2018elmo}, and BERT \citep{devlin2019bert}. 
Empirically, Long Short Term Memory (LSTM; \citet{hochreiter1997lstm}), Gated Recurrent Units (GRU; \citet{cho2014learning}), and convolutional neural networks (CNN; \citet{lecun1989cnn}) have achieved improved performance compared to other supervised machine learning models based on shallow features \citep{ren2016improving}.

Classifying SCs of scientific documents is usually based on metadata, since full text is not available for most papers and processing a large amount of full text is computationally expensive. Most existing methods for SC classification are unsupervised. For example, the Smart Local Moving Algorithm identified topics in PubMed based on text similarity \citep{boyack2018accurately} and citation information \citep{vanEck2017citation}. K-means was used to cluster articles based on semantic similarity \citep{wang2017clustering}. The \textit{memetic} algorithm, a type of evolutionary computing\citep{moscato2003gentle}, was used to classify astrophysical papers into subdomains using their citation networks. A hybrid clustering method was proposed based on a combination of bibliographic coupling and textual similarities using the \textit{Louvain} algorithm -- a greedy method that extracted communities from large networks \citep{glanzel2017using}. Another study constructed a publication-based classification system of science using the WoS dataset \citep{waltman2012new}. The clustering algorithm, described as a modularity-based clustering, is conceptually similar to $k$-nearest neighbor ($k$NN). It starts with a small set of seed labeled publications and grows by incrementally absorbing similar articles using co-citation and bibliographic coupling. Many methods mentioned above rely on citation relationships. Although such information can be manually obtained from large search engines such as Google Scholar, it is non-trivial to scale this for millions of papers. 

Our model classifies papers based only on abstracts, which are often available. Our end-to-end system is trained on a large number of labeled data with no references to external knowledge bases. When compared with citation-based clustering methods, we believe it to be more scalable and portable. 

\section{Text Representations}
\label{wordembedding}
For this work, we represent each abstract using a BoW model weighted by TF-IDF. However, instead of building a sparse vector for all tokens in the vocabulary, we choose word tokens with the highest TF-IDF values and encode them using WE models. We explore both pre-trained and re-trained WE models. We also explore their effect on classification performance based on token order. As evaluation baselines, we compare our best model with off-the-shelf text embedding models, such as the Unified Sentence Encoder (USE; \citet{cer2018universal}). We show that our model which uses the traditional and relatively simple BoW representation is computationally less expensive and can be used to classify scholarly papers at scale, such as those in the CiteSeerX repository \citep{giles1998jcdl,wu2014iaai}. 
\vspace{10pt}
\subsection{Representing Abstracts}
First, an abstract is tokenized with white spaces, punctuation, and stop words removed. Then a list $\mathbf{A}$ of word types (unique words) $w_{i}$ is generated after Wordnet lemmatization which uses the WordNet database \citep{Fellbaum2005-FELWAW} for the lemmas. 
\begin{equation}\label{eq:1}
\mathbf{A} = [w_1, w_2, w_3\dots w_n].
\end{equation}

Next the list $\mathbf{A}_f$ is sorted in descending order by TF-IDF giving $\mathbf{A}_{\rm sorted}$. TF is the term frequency in an abstract and IDF is the inverse document frequency calculated using the number of abstracts containing a token in the entire WoS abstract corpus with now: 
\begin{equation}\label{eq:3}
\mathbf{A}_{\rm sorted} = [w^{\prime}_{1}, w^{\prime}_{2}, w^{\prime}_{3}\dots w^{\prime}_{n}].
\end{equation}

Because abstracts have different numbers of words, we chose the top $d$ elements from $\mathbf{A}_{\rm sorted}$ to represent the abstract. We then re-organize the elements according to their original order in the abstract forming a sequential input. 
If the number of words is less than $d$, we pad the feature list with zeros. The final list is a vector built by concatenating all word level vectors $\vec{\mathbf{v}}^{\prime}_{k},k\in \{1,\cdots,d\}$ into a $D_{\rm WE}$ dimension vector. The final semantic feature vector $\mathbf{A}_f$ is:
\begin{equation}\label{eq:4}
\mathbf{A}_f = [\vec{\mathbf{v}}^{\prime}_{1}, \vec{\mathbf{v}}^{\prime}_{2}, \vec{\mathbf{v}}^{\prime}_{3}\dots \vec{\mathbf{v}}^{\prime}_{d}]
\end{equation}

\subsection{Word Embedding}
To investigate how different word embeddings affect classification results, we apply several widely used models. An exhaustive experiment for all possible models is beyond the scope of this paper. We use some of the more popular ones as now discussed.

\textbf{GloVe} captures semantic correlations between words using global word-word co-occurrence, as opposed to local information used in word2vec \citep{mikolov2013word2vec}. It learns a word-word co-occurrence matrix and predicts co-occurrence ratios of given words in context \citep{pennington2014glove}. Glove is a context-independent model and outperformed other word embedding models such as word2vec in tasks such as word analogy, word similarity, and named entity recognition tasks.

\textbf{FastText} is another context-independent model which uses sub-word (e.g., character $n$-grams) information to represent words as vectors \citep{bojanowski2017enriching}. It uses log-bilinear models that ignore the morphological forms by assigning distinct vectors for each word. 
If we consider a word $w$ whose $n$-grams are denoted by $g_w$, then the vector $\mathbf{z}_g$ is assigned to each $n$-grams in $g_w$. 
Each word is represented by the sum of the vector representations of its character $n$-grams. This representation is incorporated into a Skip Gram model \citep{goldberg2014word2vec} which improves vector representation for morphologically rich languages.

\textbf{SciBERT} is a variant of BERT, a context-aware WE model that has improved the performance of many NLP tasks such as question answering and inference \citep{devlin2019bert}. The bidirectionally trained model seems to learn a deeper sense of language than single directional transformers. 
The transformer uses an attention mechanism that learns contextual relationships between words.  
{SciBERT} uses the same training method as BERT but is trained on research papers from Semantic Scholar. Since the abstracts from WoS articles mostly contain scientific information, we use SciBERT \citep{beltagy2019scibert} instead of BERT. Since it is computationally expensive to train BERT (4 days on 4--16 Cloud TPUs as reported by Google), we use the pre-trained SciBERT.

\subsection{Retrained WE Models}
Though pretrained WE models represent richer semantic information compared with traditional one-hot vector methods, when applied to text in scientific articles the classifier does not perform well. This is probably because the text corpus used to train these models are mostly from Wikipedia and Newswire. The majority of words and phrases included in the vocabulary extracted from these articles provides \textit{general} descriptions of knowledge, which are significantly different from those used in scholarly articles which describe specific \textit{domain} knowledge. Statistically, the overlap between the vocabulary of pretrained GloVe (6 billion tokens) and WoS is only 37\% \citep{wu2018bigdata}. Nearly all of the WE models can be retrained. Thus, we retrained GloVe and FastText using 6.38 million abstracts in WoS (by imposing a limit of 150K on each SC, see below for more details). The token size obtained after training WoS abstracts is 1.13 billion with sizes of 1 million for GloVe and 1.2 million for FastText.

\subsection{Universal Sentence Encoder}
For baselines, we compared with Google's Universal Sentence Encoder (USE)  and the character-level convolutional network (CCNN). USE uses transfer learning to encode sentences into vectors. The architecture consists of a transformer-based sentence encoding\citep{vaswani2017attention} and a deep averaging network (DAN)\citep{iyyer2015deep}. These two variants have trade-offs between accuracy and compute resources. We chose the transformer model because it performs better than the DAN model on various NLP tasks \citep{cer2018universal}. CCNN is a combination of character-level features trained on temporal (1D) convolutional networks (ConvNets; \citet{zhang2015nips}). It treats input characters in text as a raw-signal which is then applied to ConvNets. Each character in text is encoded using a one-hot vector such that the maximum length $l$ of a character sequence does not exceed a preset length $l_0$. 

\begin{figure*}[t]
\centering
\includegraphics[width=\textwidth]{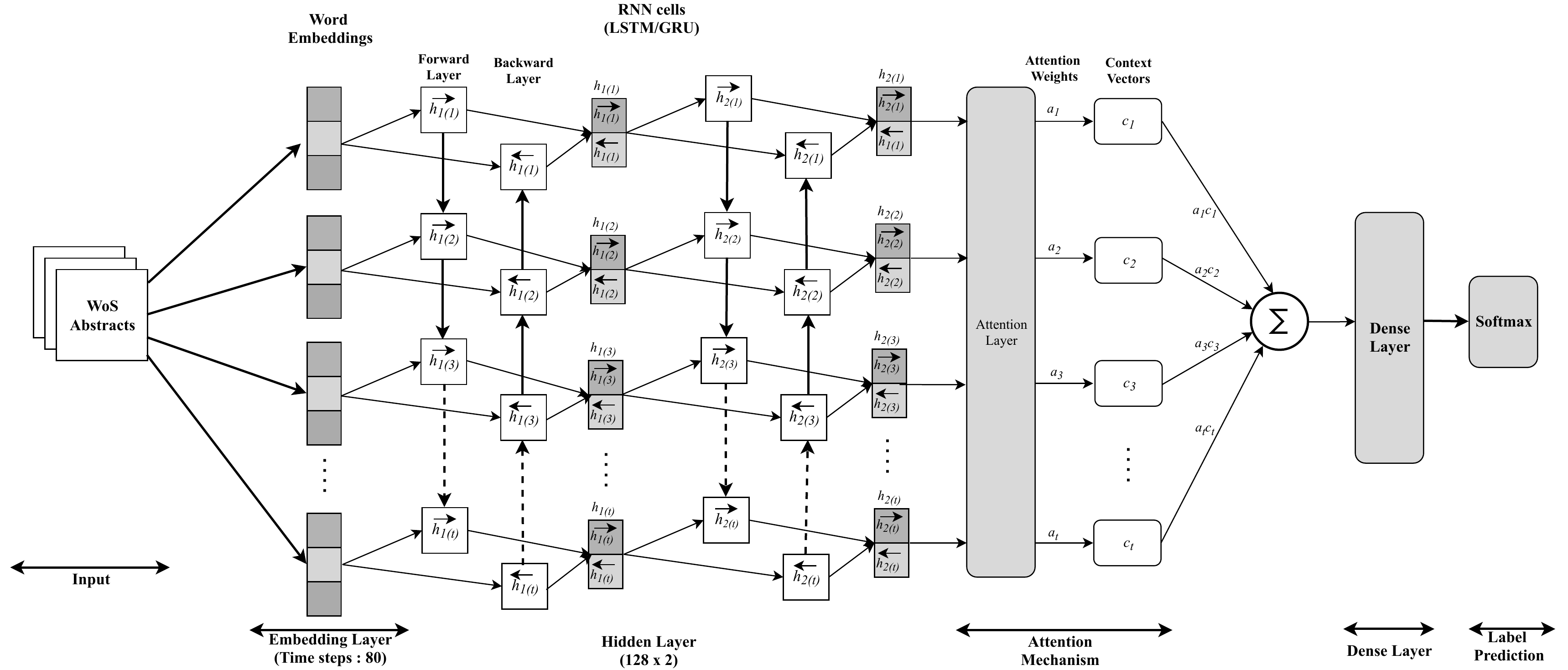}
\caption{Subject category (SC) classification architecture.}\label{architecture}
\end{figure*}

\section{Classifier Design}
The architecture of our proposed classifier is shown in Figure~\ref{architecture}. An abstract representation previously discussed is passed to the neural network for encoding. Then the label of the abstract is determined by the output of the sigmoid function that aggregates all word encodings. Note that this architecture is not applicable for use by CCNN or USE. For comparison, we used these two architectures directly as described from their original publications.  

{\bf LSTM} 
is known for handling the vanishing gradient that occurs when training recurrent neural networks. A typical LSTM cell consists of 3 gates: input gate $\mathbf{i}_t$, output gate $\mathbf{o}_t$ and forget gate $\mathbf{f}_t$. The input gate updates the cell state; the output gate decides the next hidden state, and the forget gate decides whether to store or erase particular information in the current state $\mathbf{h}_t$. We use $\tanh(\cdot)$ as the activation function and the sigmoid function $\sigma(\cdot)$ to map the output values into a probability distribution.
The current hidden state $\mathbf{h}_t$ of LSTM cells can be implemented with the following equations:
\begin{equation}\label{eq:lstm_1}
\mathbf{i}_t = \sigma(\mathbf{W}_i\mathbf{x}_t+\mathbf{U}_i\mathbf{h}_{t-1}+\mathbf{b}_i)
\end{equation}
\begin{equation}\label{eq:lstm_2}
\mathbf{f}_t = \sigma(\mathbf{W}_f\mathbf{x}_t+\mathbf{U}_f\mathbf{h}_{t-1}+\mathbf{b}_f)
\end{equation}
\begin{equation}\label{eq:lstm_3}
\mathbf{z}_t = \tanh(\mathbf{W}_z\mathbf{x}_t+\mathbf{U}_z\mathbf{h}_{t-1}+\mathbf{b}_z)
\end{equation}
\begin{equation}\label{eq:lstm_4}
\mathbf{c}_t = \mathbf{z}_t \odot \mathbf{i}_t + \mathbf{c}_{t-1} \odot \mathbf{f}_t
\end{equation}
\begin{equation}\label{eq:lstm_5}
\mathbf{o}_t = \sigma(\mathbf{W}_o\mathbf{x}_t+\mathbf{U}_o\mathbf{h}_{t-1}+\mathbf{b}_o)
\end{equation}
\begin{equation}\label{eq:lstm_6}
\mathbf{h}_t = \tanh(\mathbf{c}_t) \odot \mathbf{o}_t
\end{equation}

At a given time step $t$,  $\mathbf{x}_t$ represents the input vector; $\mathbf{c}_t$ represents cell state vector or memory cell;  $\mathbf{z}_t$ is a temporary result. $\mathbf{W}$ and $\mathbf{U}$ are weights for the input gate $i$, forget gate $f$, temporary result $z$, and output gate $o$.

{\bf GRU} is similar to LSTM,
except that it has only a reset gate $\mathbf{r}_t$ and an update gate $\mathbf{z}_t$. The current hidden state $\mathbf{h}_t$ at a given timestep $t$ can be calculated with: 
\begin{equation}\label{eq:gru_1}
\mathbf{z}_t = \sigma(\mathbf{W}_z\mathbf{x}_t+\mathbf{U}_z\mathbf{h}_{t-1}+\mathbf{b}_z)
\end{equation}
\begin{equation}\label{eq:gru_2}
\mathbf{r}_t = \sigma(\mathbf{W}_r\mathbf{x}_t+\mathbf{U}_r\mathbf{h}_{t-1}+\mathbf{b}_r)
\end{equation}
\begin{equation}\label{eq:gru_3}
\mathbf{\tilde{h}}_t = \tanh(\mathbf{W}_h\mathbf{x}_t+\mathbf{U}_h(\mathbf{r}_t \odot \mathbf{h}_{t-1})+\mathbf{b}_h)
\end{equation}
\begin{equation}\label{eq:gru_4}
\mathbf{h}_t = (1-\mathbf{z}_t) \odot \mathbf{h}_{t-1} + \mathbf{z}_t \odot \mathbf{\tilde{h}}_t
\end{equation} 
with the same defined variables. GRU is less computationally expensive than LSTM and achieves comparable or better performance for many tasks. For a given sequence, we train LSTM and GRU in two directions (BiLSTM and BiGRU) to predict the label for the current position using both historical and future data, which has been shown to outperform a single direction model for many tasks. 

\begin{figure*}[b]
\centering
\includegraphics[width=\textwidth]{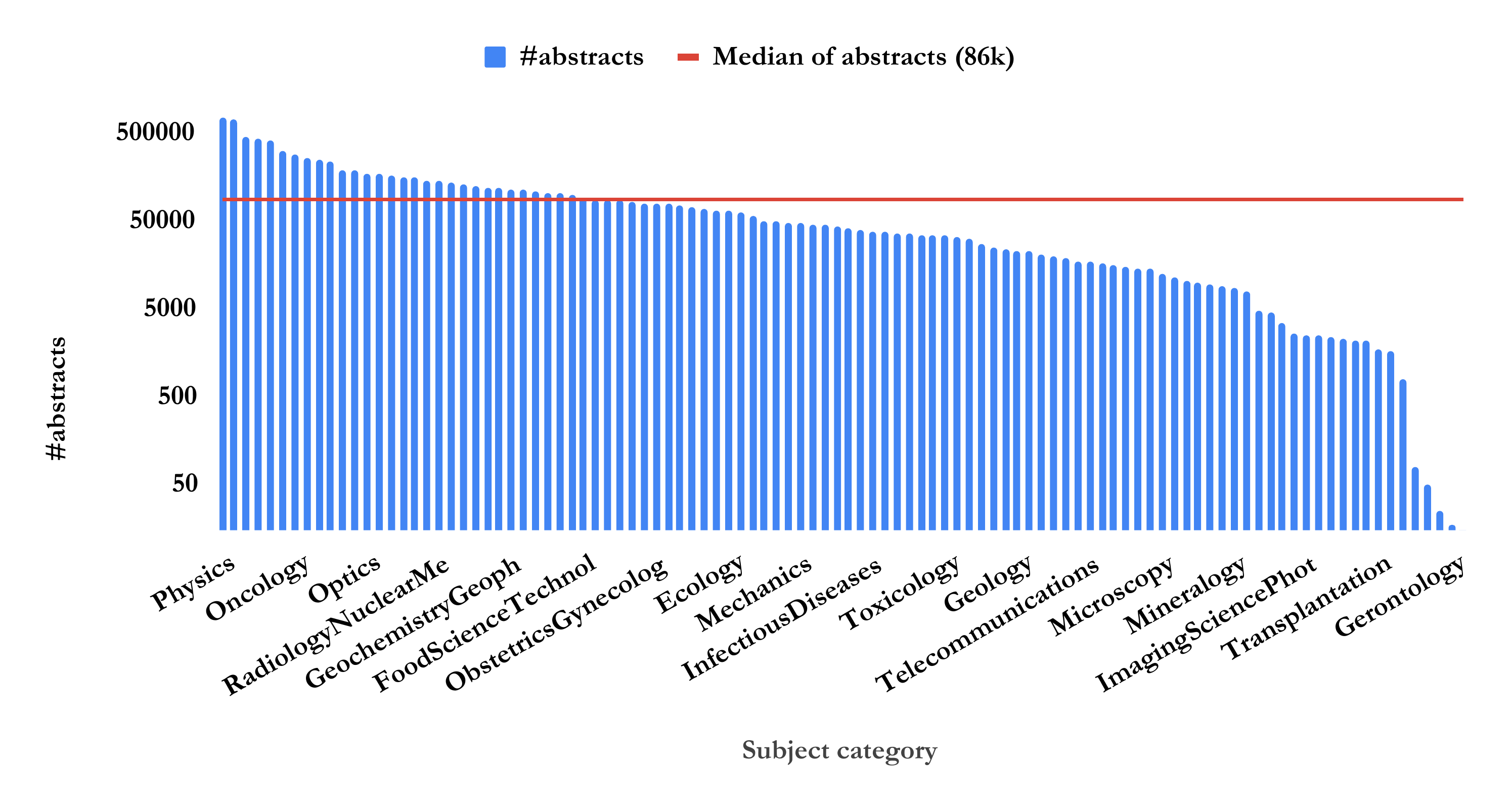}
\caption{Distribution of numbers of abstracts in the 104 SCs for our corpus. $y$-axis is logarithmic. Red line marks the median number of abstracts.}\label{fig:scdist}
\end{figure*}

{\bf Attention Mechanism} The attention mechanism is used to weight word tokens deferentially when aggregating them into a document level representations. In our system (Figure~\ref{architecture}), embeddings of words are concatenated into a vector with $D_{\rm WE}$ dimensions. 
Using the attention mechanism, each word $t$ contributes to the sentence vector, which is characterized by the factor $\alpha_t$ such that
\begin{equation}
\alpha_t=\frac{\exp\left({\mathbf{u}_t^\top\mathbf{v}_t}\right)}{\sum_t\exp\left({\mathbf{u}_t^\top\mathbf{v}_t},\right)}
\end{equation}
\begin{equation}
    \quad \mathbf{u}_t=\tanh\left(\mathbf{W}\cdot\mathbf{h}_t+\mathbf{b}\right)
\end{equation}
in which $\mathbf{h}_t=[\overrightarrow{\mathbf{h}}_t;\overleftarrow{\mathbf{h}}_t]$ is the representation of each word after the BiLSTM or BiGRU layers, $\mathbf{v}_t$ is the context vector that is randomly initialized and learned during the training process, $\mathbf{W}$ is the weight, and $\mathbf{b}$ is the bias. An abstract vector $\mathbf{v}$ is generated by aggregating word vectors using weights learned by the attention mechanism. We then calculate the weighted sum of $\mathbf{h}_t$ using the attention weights by:
\begin{equation}
\mathbf{v}=\sum_t\alpha_t\mathbf{h}_t.
\end{equation}

\section{Experiments}
Our training dataset is from the WoS database for the year 2015. The entire dataset contains approximately 45 million academic documents, nearly all of which have titles and abstracts from published papers. They are labeled with 235 SCs in three broad categories--Science, Social Science, and Art and Literature. A portion of the SCs have subcategories, such as ``Physics, Condensed Matter'', ``Physics, Nuclear'', and ``Physics, Applied''. 
Here, we collapse these subcategories, which reduces the total number of SCs to 115. We do this because the minor classes decrease the performance of the model (due to the less availability of that data). 
Also, we need to have an "others" class to balance the data samples. 
We also exclude papers labeled with more than one category and papers that are labeled as ``Multidisciplinary''. Abstracts with less than 10 words are excluded. The final number of singly labeled abstracts is approximately 9 million, labeled with 104 SCs. Figure~\ref{fig:scdist} illustrates the distribution of abstracts across SCs which indicates that the labelled dataset contains imbalanced sample sizes of SCs ranging from 15 (Art) to 734K (Physics). 
We randomly select up to 150K abstracts per SC. This upper limit is based on our preliminary study \citep{wu2018bigdata}. The ratio between the training and testing corpus is 9:1.

The median of word types per abstract is approximately 80 to 90. As such, we choose the top $d=80$ elements from $\mathbf{A}_{\rm sorted}$ to represent the abstract. If $\mathbf{A}_{\rm sorted}$ has less than $d$ elements, we pad the feature list with zeros. The word vector dimensions of GloVe and FastText are set to 50 and 100, respectively. This falls into the reasonable value range (24--256) for WE dimensions \citep{witt2017tpdl}. 
When training the BiLSM and BiGRU models, 
each layer contains 128 neurons. 
We investigate the dependency of classification performance on these hyper-parameters by varying the number of layers and neurons.
We varied the number of word types per abstract $d$ and set the dropout rate to 20\%  to mitigate overfitting or underfitting. Due to their relatively large size, we train the neural networks using mini-batch gradient descent with \textit{Adam} for gradient optimization and one word cross entropy as the loss function. The learning rate was set at $10^{-3}$.

We used the CCNN architecture \citep{zhang2015nips}, which contains 6 convolutional layers each including 1008 neurons followed by 3 fully-connected layers. Each abstract is represented by a 1014 dimensional vector. Our architecture for USE \citep{cer2018universal} was an MLP that contained 4 layers, each of which contained 1024 neurons. Each abstract is represented by a 512 dimensional vector. 

\begin{figure}
\includegraphics[width=0.5\textwidth]{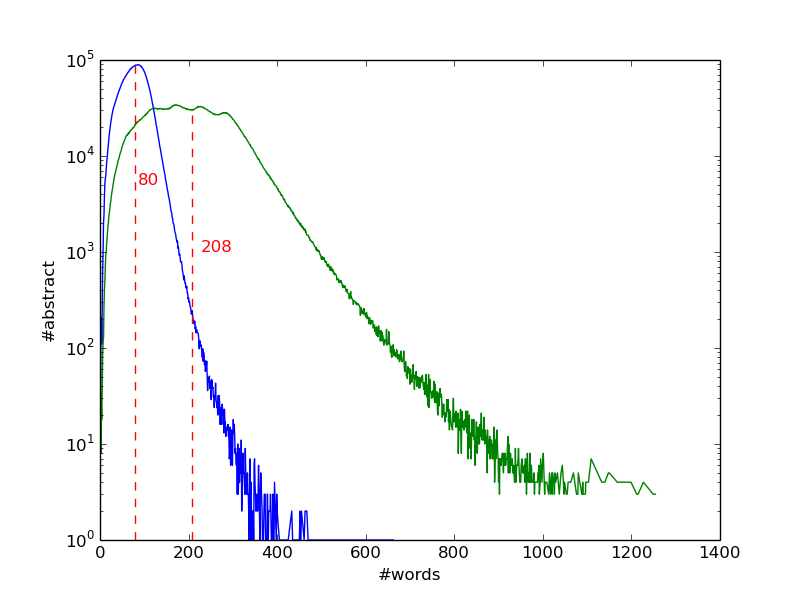}
\caption{Distribution of the number of  words in an abstract. Blue line represents the number of unique words in an abstract and green line represents the distribution of non-unique words. Red dotted lines represent the median of the words (unique and non-unique respectively). Median number of unique words is 80. }\label{words_distribution}
\end{figure}

\section{Evaluation and Comparison}

\begin{figure}[!ht]
\includegraphics[width=0.5\textwidth]{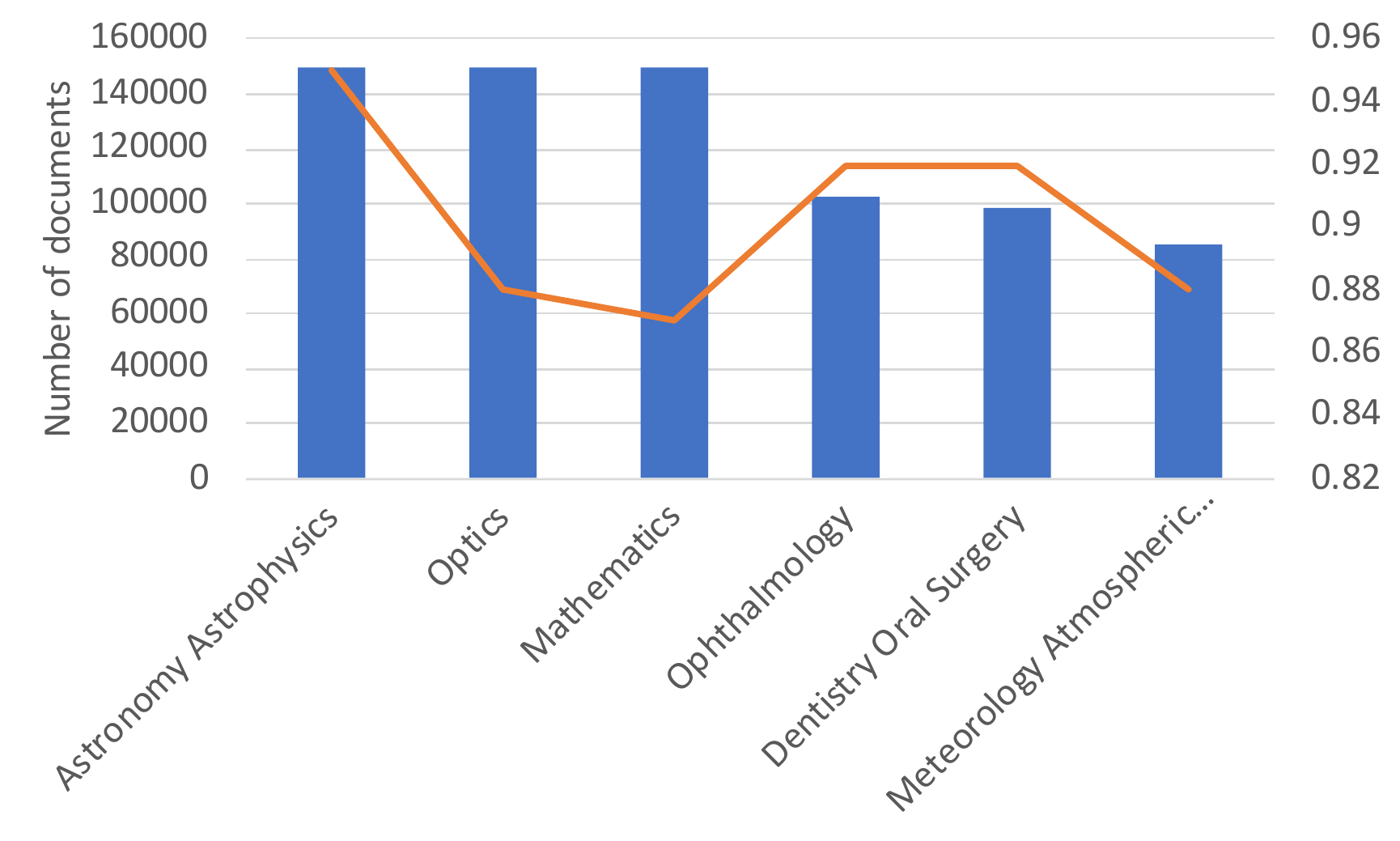}
\includegraphics[width=0.5\textwidth]{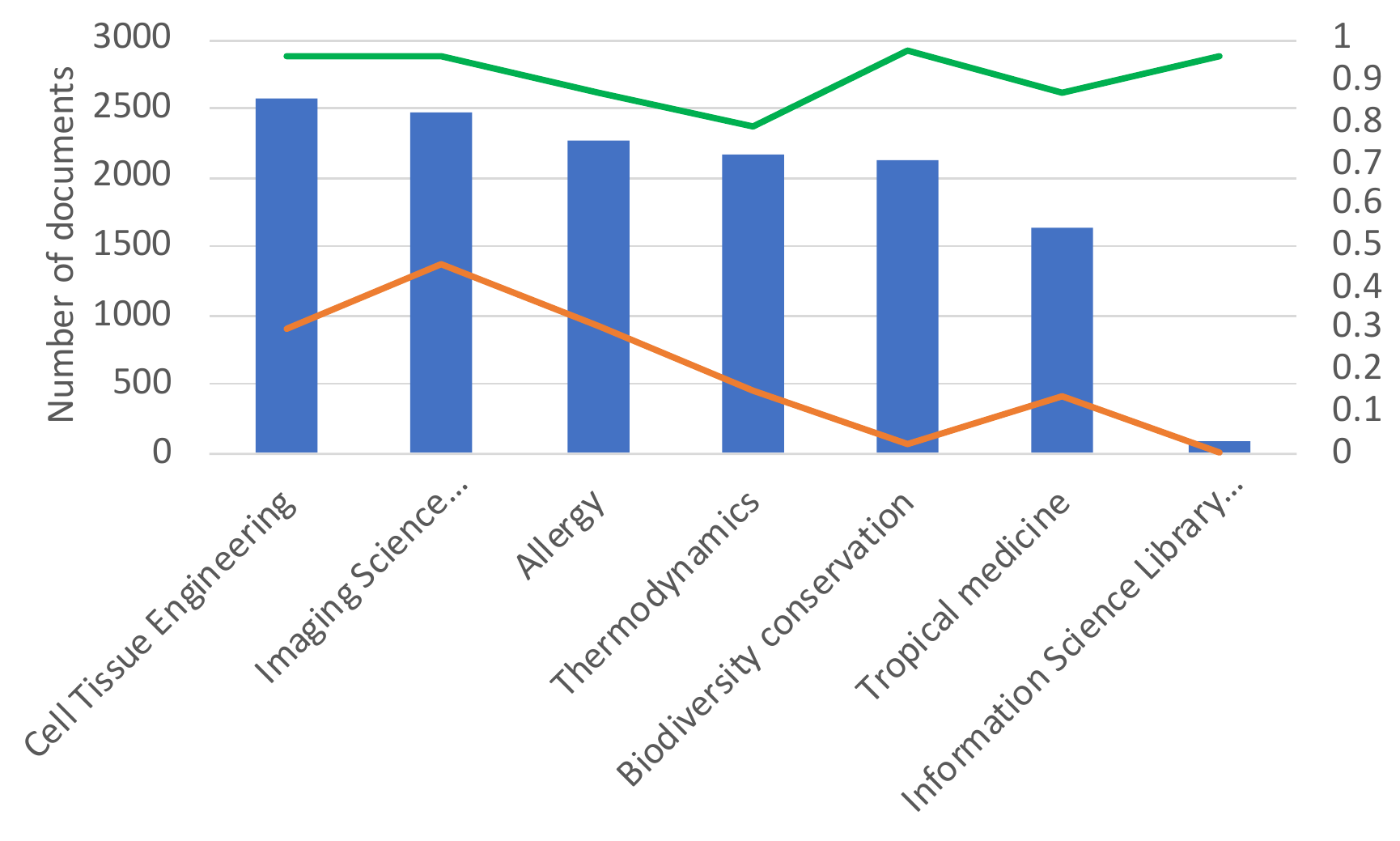}
\caption{\small Number of training documents (blue bars) and the corresponding $F_1$ values (red curves) for best performance (\emph{First}) and worst performance (\emph{Second}) SC's. Green line on the right panel shows improved $F_1$'s produced by the second-level classifier.}\label{fig:sc104}
\end{figure}

\subsection{One-level Classifier}
We first classify all abstracts in the testing set into 104 SCs using the retrained GloVe WE model with BiGRU. The model achieves a micro-$\mathrm{F_1}$ score of 0.71. The first panel in Figure~\ref{fig:sc104} shows the SCs that achieve the highest ${F_1}$'s; the second panel shows SCs that achieve relatively low ${F_1}$'s. The results indicate that the classifier performs poorer on SCs with relatively small sample sizes than SCs with relatively large sample sizes. The data imbalance is likely to contribute to the significantly different performances of SCs. \\

\begin{figure*}
        \centering
        \includegraphics[width=\textwidth]{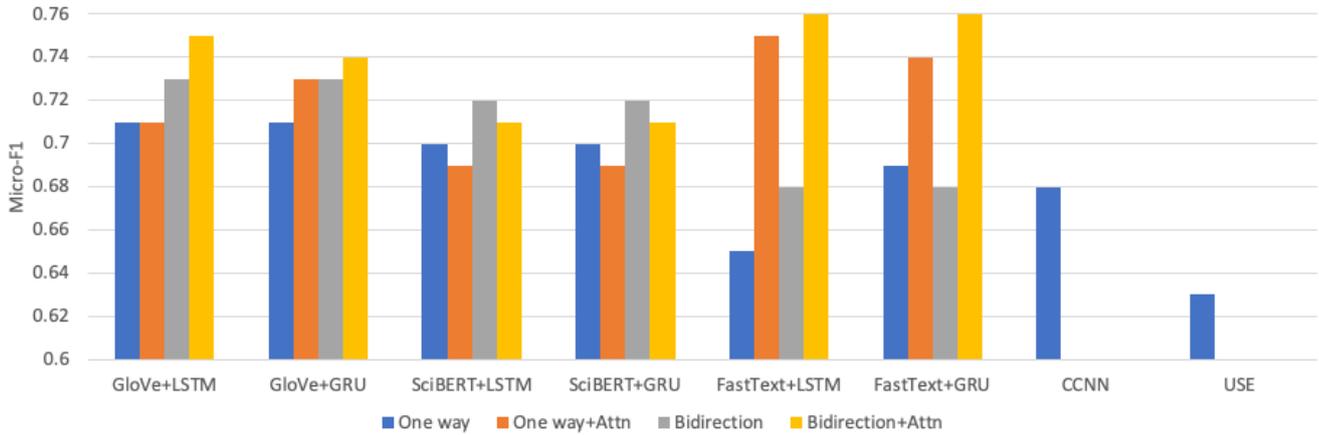}
        \caption{Micro-$F_1$'s of models that classify abstracts into 81 SCs. 
        Variants of models within each group are color-coded. CCNN hyper-parameters are unalterable. }
        \label{fig:sc81results}
\end{figure*}
\vspace{10pt}

\subsection{Two-level Classifier}
\begin{figure*}[t]
\centering
\captionsetup{justification=centering}
\includegraphics[width=\textwidth]{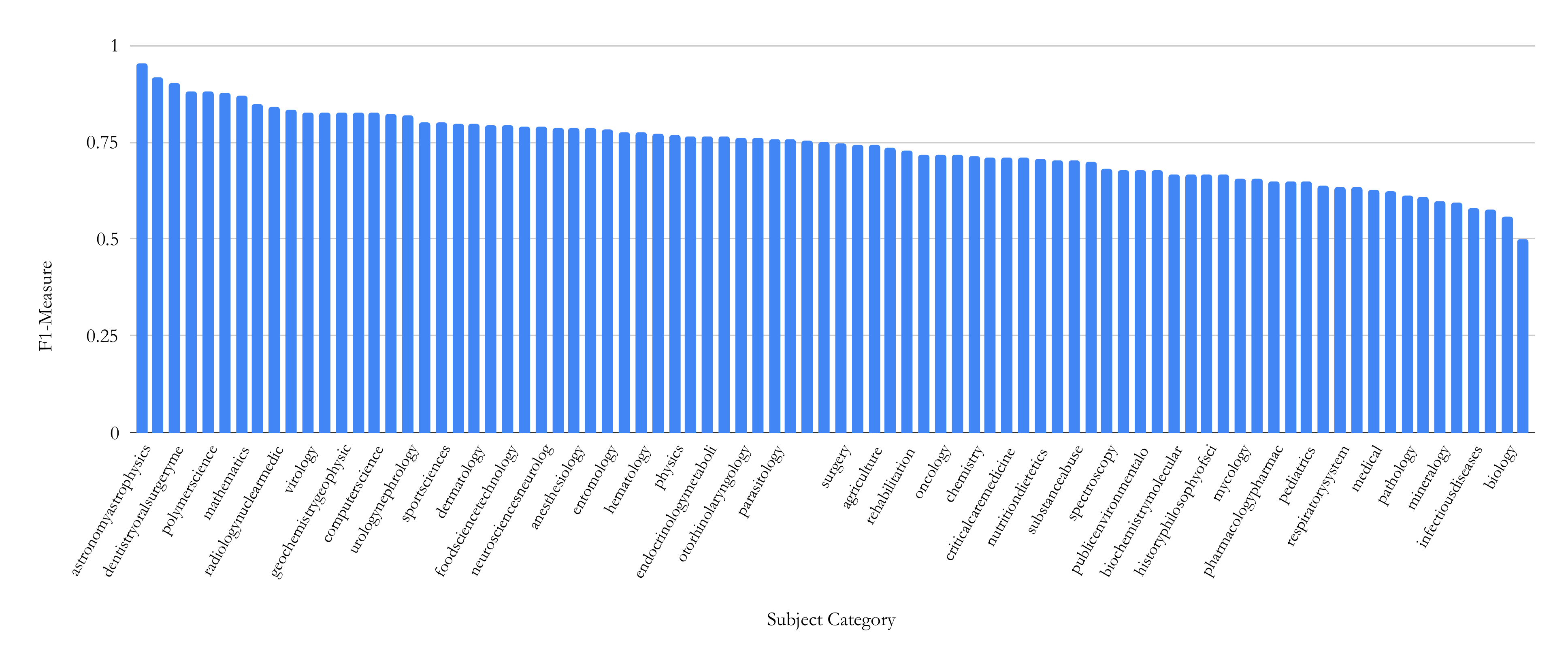}
\caption{Distribution of {$F_1$}'s across 81 SC's obtained by the first layer classifier.}\label{subject_f1_scores}
\end{figure*}

To mitigate the data imbalance problems for the one-level classifier, we train a two-level classifier. The first level classifies abstracts into 81 SCs, including 80 major SCs and an ``Others" category, which incorporates 24 minor SCs. ``Others'' contains the categories with training data $<$ 10k abstracts. Abstracts that fall into the ``Others'' are further classified by a second level classifier, which is trained on abstracts belonging to the 24 minor SCs. Below are observations based on the results in Figure~\ref{fig:sc81results}. Please refer to supplementary material for the comprehensive results using the two-level classifier. 
\begin{enumerate}[nolistsep,leftmargin=*]
    \item FastText+BiGRU+Attn and FastText+\\BiLSTM+Attn achieve the highest micro-$F_1$ of 0.76. Several models achieve similar results:\\
    GloVe+BiLSTM+Attn (micro-$F_1$=0.75), \\
    GloVe+BiGRU+Attn (micro-$F_1$=0.74), \\
    FastText+LSTM+Attn (micro-$F_1$=0.75), \\
    and FastText+GRU+Attn (micro-$F_1$=0.74). These results indicate that the attention mechanism significantly improves the classifier performance.
    \item The best performance of FastText and GloVe can be partially attributed to the retrained WE. In contrast, the best micro-${F_1}$ achieved by USE is 0.64, which is likely resulted from its relatively low vocabulary overlap. Another reason could be is that the single vector of fixed length only encodes the overall semantics of the abstract. The occurrences of words are better indicators of sentences in specific domains. 
    \item LSTM and GRU and their counterparts exhibit very similar performance, which is consistent with a recent systematic survey \citep{greff2017lstm}.
    \item For FastText+BiGRU+Attn, the $F_1$ measure varies from {0.50} to {0.95} with a median of {0.76}. The distribution of $F_1$ values for 81 SCs is shown in Figure~\ref{subject_f1_scores}. The $F_1$ achieved by the first-level classifier with 81 categories (micro-$F_1$=0.76) is improved compared with the classifier trained on 104 SCs (micro-$F_1$=0.70) 
    \item The performance was not improved by increasing the GloVe vector dimension from 50 to 100 (not shown) under the setting of GloVe+BiGRU with 128 neurons on 2 layers which is consistent with earlier work \citep{witt2017tpdl}.
    \item Word-level embedding models in general perform better than the character-level embedding models (i.e., CCNN). CCNN considers the text as a raw-signal, so the word vectors constructed are more appropriate when comparing morphological similarities. However, semantically similar words may not be morphologically similar, e.g., ``Neural Networks'' and ``Deep Learning''.
    \item SciBERT's performance is 3-5\% below FastText and GloVe, indicating that re-trained WE models exhibit an advantage over pre-trained WE models. This is because SciBERT was trained on the PubMed corpus which mostly incorporates papers in biomedical and life sciences. Also, due to their large dimensions, the training time was greater than FastText under the same parameter settings.
\end{enumerate}

We also investigated the dependence of classification performance on key hyper-parameters. The settings of GLoVe+BiGRU with 128 neurons on 2 layers are considered as the ``reference setting''. With the setting of GloVe+BiGRU, we increase the neuron number by factor of 10 (1280 neurons on 2 layers) and obtained marginally improved performance by 1\% compared with the same setting with 128 neurons. We also double the number of layers (128 neurons on 4 layers). Without attention, the model performs worse than the reference setting by 3\%. With the attention mechanism, the micro-$F_1=0.75$ is marginally improved by 1\% with respect to the reference setting. We also increase the default number of neurons of USE by 2, making it 2048 neurons for 4 layers. The micro-$F_1$ improves marginally by 1\%, reaching only $0.64$. The results indicate that adding more neurons and layers seem to have little impact to the performance improvement.

The second-level classifier is trained using the same neural architecture as the first-level on the ``Others'' corpus. 
Figure~\ref{fig:sc104} (\emph{Right ordinate legend}) shows that ${F_1}$'s vary from 0.92 to 0.97 with a median of 0.96. The results are significantly improved by classifying minor classes separately from major classes.



\section{Discussion}
\begin{figure*}[t]%
    \centering
    \includegraphics[width=\textwidth]{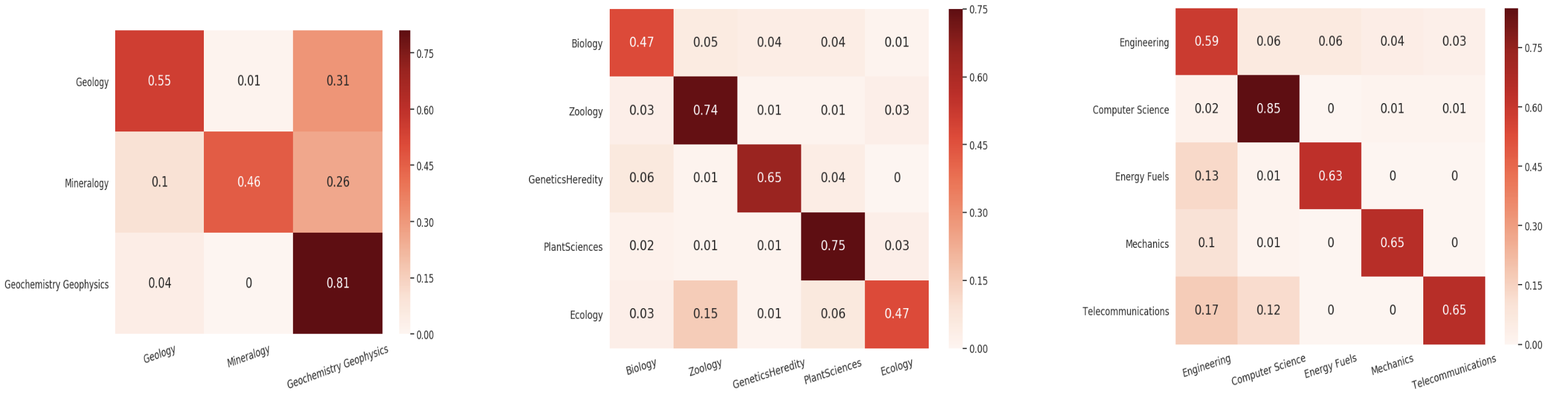}
    \caption{Normalized Confusion Matrix for closely-related classes in which a large fraction of ``Geology" and ``Mineralogy" papers are classified into ``GeoChemistry GeoPhysics (left), and a large fraction of Zoology papers are classified into "biology" or "ecology" (middle), a large fraction of ``TeleCommunications",``Mechanics'' and ``EnergyFuels" papers are classified into ``Engineering"(right). }%
    \label{fig:closely-related}%
\end{figure*}

\subsection{Sampling Strategies}
The data imbalance problem is ubiquitous in both multi-class and multi-label classification problems \citep{charte2015addressing}. The imbalance ratio (IR), defined as the ratio of the number of instances in the majority class to the number of samples in the minority class \citep{garcia2012effectiveness}, has been commonly used to characterize the level of imbalance. Compared with the imbalance datasets in Table~1 of \citep{charte2015addressing}, our data has a significantly high level of imbalance. In particular, the highest IR is about 49,000 (\#Physics/\#Art).  
One commonly used way to mitigate this problem is data resampling. This method is based on rebalancing SC distributions by either deleting instances of major SCs (undersampling) or supplementing artificially generated instances of the minor SCs (oversampling). We can always undersample major SCs, but this means we have to reduce sample sizes of all SCs down to about 15 (Art; Figure~\ref{fig:scdist}), which is too small for training robust neural network models. The oversampling strategies such as SMOTE \citep{chawla2002smote} works for problems involving continuous numerical quantities, e.g., \citet{salaheldeen2015predicting}. In our case, the synthesize vectors of ``abstracts'' by SMOTE will not map to any actual words because word representations are very sparsely distributed in the large WE space. Even if we oversample minor SCs using these semantically dummy vectors, generating all samples will take a large amount of time given the high dimensionality of abstract vectors and high IR. Therefore, we only use our real data.

\begin{figure*}[t]
\centering
\includegraphics[width=\textwidth]{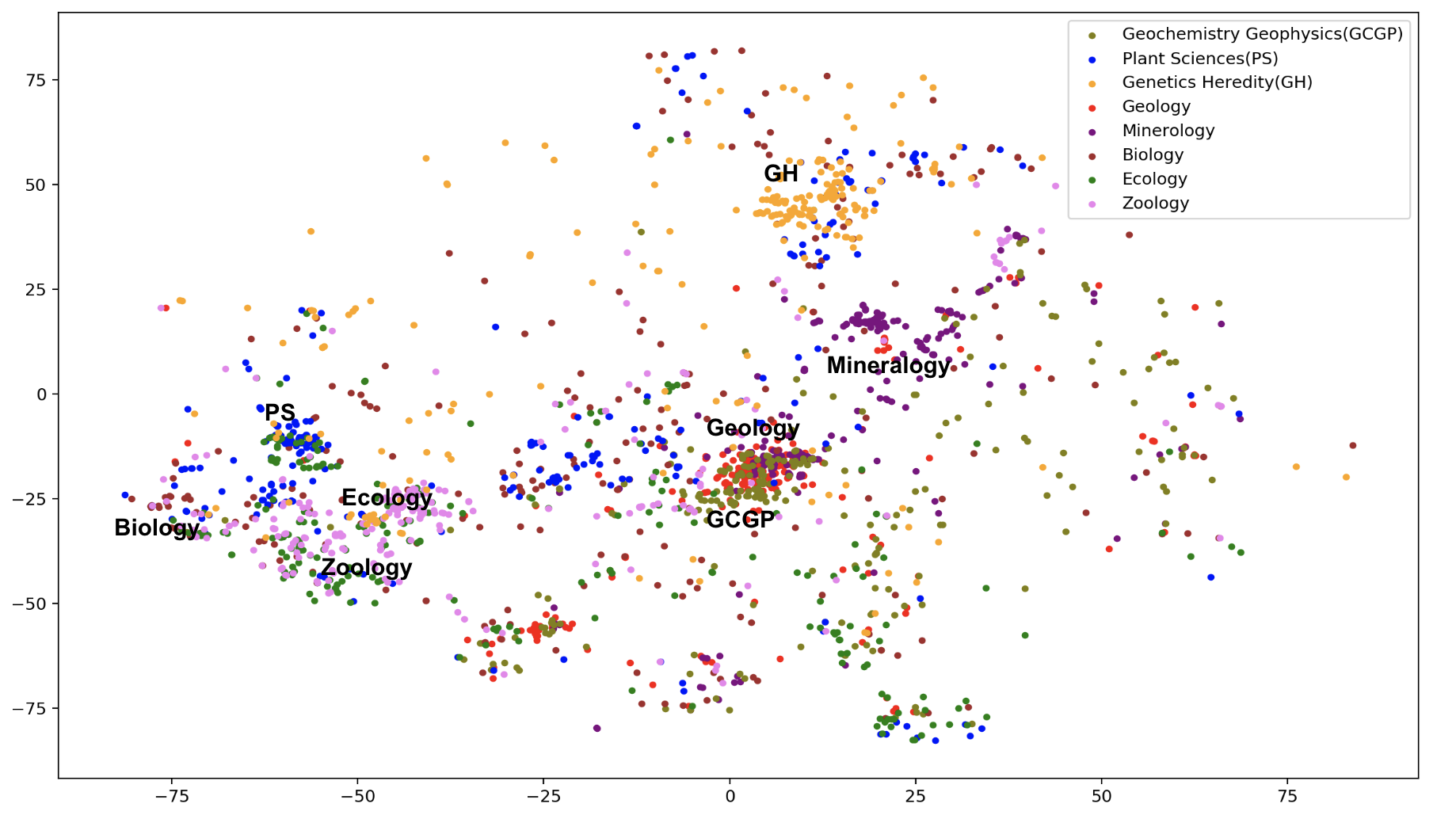}
\caption{t-SNE plot of closely-related SCs.}\label{fig:tsne_closely_related}
\end{figure*}
\vspace{6pt}

\subsection{Category Overlapping}
We discuss the potential impact on classification results contributed by categories overlapping in the training data. 
Our initial classification schema contains 104 SCs, but they are not all mutually exclusive. Instead, the vocabularies of some categories overlap with the others. 
For example, papers exclusively labeled as ``Materials Science" and ``Metallurgy" exhibit significant overlap in their tokens. 
In the WE space, the semantic vectors labeled with either category are overlapped making it hard to differentiate them. Figure~\ref{fig:closely-related} shows the confusion matrices of the closely related categories such as ``Geology", ``Mineralogy", and ``Geochemistry Geophysics''. Figure~\ref{fig:tsne_closely_related} is the t-SNE plot of abstracts of closely related SCs. To make the plot less crowded, we randomly select 250 abstracts from each SC as shown in Figure~\ref{fig:closely-related}. Data points representing  ``Geology", ``Mineralogy", and ``Geochemistry Geophysics" tend to spread or are overlapped in such a way that are hard to be visually distinguished. 


One way to mitigate this problem is to merge overlapped categories. However, special care should be taken on whether these overlapped SCs are truly strongly related and should be evaluated by domain experts.  
For example, 
``Zoology'', ``PlantSciences'', and ``Ecology'' \emph{can} be merged into a single SC called ``Biology'' (Gaff 2019, private communication).  
``Geology'', ``Mineralogy'', and ``GeoChemistry GeoPhysics'' can be merged into a single SC called ``Geology''. 
However, ``Materials Science'' and ``Metallurgy'' may \emph{not} be merged (Liu 2019, private communication) to a single SC. 
By doing the aforementioned merges, the number of SCs is reduced to 74. As a preliminary study, we classified the merged dataset using our best model (retrained FastText+BiGRU+Attn) and achieved an improvement with an overall micro-${F_1}$ score of 0.78. The ``Geology'' SC after merging has significantly improved from $0.83$ to $0.88$.



\vspace{8pt}

\subsection{Application to CiteSeerX}
\begin{figure*}[t]
    \centering
    \includegraphics[width=.8\textwidth]{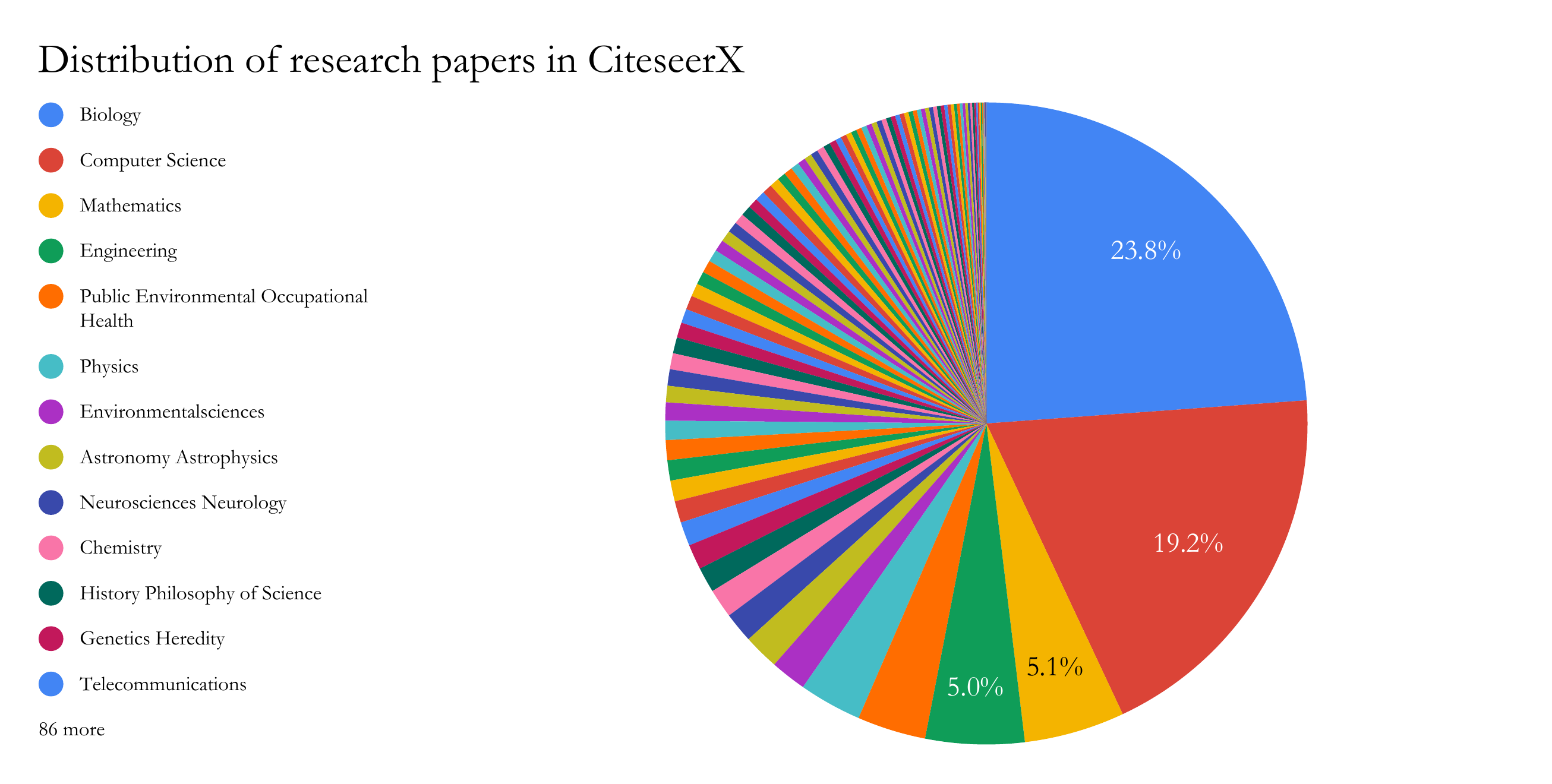}
    \caption{Classification results of 1 million research papers in CiteSeerX, using our best model.}
    \label{fig:csxclass}
\end{figure*}

CiteSeerX is a digital library search engine that was the first to use automatic citation indexing\citep{giles1998jcdl}. It is an open source search engine that provides metadata and full-text access for more than 10 million scholarly documents and continues to add new documents \citep{wu2019citeseerx}. 
In the past decade, it has incorporated scholarly documents in diverse SCs, but the distribution of their subject categories is unknown.  
Using the best neural network model in this work (FreeText+BiGRU+Attn), we classified 1 million papers randomly selected from CiteSeerX into 104 SCs (Figure~\ref{fig:csxclass}). The top five subject categories are Biology (23.8\%), Computer Science (19.2\%), Mathematics (5.1\%), Engineering (5.0\%), Public Environmental Occupational Health (3.5\%).
The fraction of Computer Science papers is significantly higher than the results in \citet{wu2018bigdata}. 
The $F_1$ for Computer Science was about 94\% which is higher than this work (about 80\%). Therefore, the fraction may be overestimated here. However, \citep{wu2018bigdata} had only 6 classes and this model classifies abstracts into 104 SCs, so although this compromises the accuracy, (by around 7\% on average), our work can still be used as a starting point for a systematic subject category classification. The classifier classifies 1 million abstracts in 1253 seconds implying that will be scalable on multi-millions of papers.

\section{Conclusions}
We investigated the problem of systematically classifying a large collection of scholarly papers into 104 SC's using neural network methods based only on abstracts. Our methods appear to scale better than existing clustering-based methods which rely on citation networks.
For neural network methods, our retrained FastText or GloVe combined with BiGRU or BiLSTM with the attention mechanism gives the best results. Retraining WE models and using an attention mechanism play important roles in improving the classifier performance. A two-level classifier effectively improves our performance when dealing with training data that has extremely imbalanced categories. The median ${F_1}$'s under the best settings are 0.75--0.76.

One bottleneck of our classifier is the overlapping categories. Merging closely related SCs is a promising solution, but should be under the guidance of domain experts. 
The TF-IDF representation only considers unigrams. Future work could consider $n$-grams ($n\geq2$) and transfer learning to adopt word/sentence embedding models trained on non-scholarly corpora \citep{conneau2017emnlp,arora2017iclr}. \citep{zhang2016improving}. . 
One could investigate models that also take into account stop-words, e.g., \citep{yang2016han}. One could also explore alternative optimizers of neural networks besides \emph{Adam}, such as the Stochastic Gradient Descent (SGD). 

\section*{Acknowledgements}
We gratefully acknowledge partial support from the National Science Foundation. We also acknowledge Adam T. McMillen for technical support, and Holly Gaff, Old Dominion University and Shimin Liu, Pennsylvania State University as domain experts respectively in biology and the earth and mineral sciences.

\bibliographystyle{frontiersBib} 
\end{document}


\onecolumn
\firstpage{1}

\title[Supplementary Material]{{\helveticaitalic{Supplementary Material}}}

\maketitle

\section{Supplementary Data}
\begin{table*}[!h]
\centering\small\caption{Comparison of models used in the first-level classifier which classifies abstracts into 81 SCs. For column $\mathbf{D}_{\rm WE}$ the integer in the parentheses is the dimension of the retrained word vectors. \textbf{\#Nr} is the number of neurons per layer. \textbf{\#Lr} is the number of hidden layers. mi$\mathbf{F_1}$ stands for micro-$F_1$. $\mathbf{T}_{\rm t}$ is the time elapsed in seconds to classify 638k abstracts in the testing corpus. CCNN and USE have fixed dimensions. }\label{tab:results_81_classes}
\begin{tabular}{c|c|lrccr} 
\toprule
\textbf{WE} & $\mathbf{D_{\rm WE}}$ & \textbf{Model} & \textbf{\#Nr} & \textbf{\#Lr} & mi$\mathbf{F_1}$ & $\mathbf{T}_{\rm t}$ \\ 
\toprule
\multirow{12}{*}{GloVe}  & \multirow{12}{*}{\begin{tabular}[c]{@{}c@{}}4000\\($50$)\end{tabular}}  & LSTM & 128 & 2 & 0.71 & 620 \\
& & LSTM + Attn     & 128   & 2 & 0.72  & 689\\ 
\cmidrule{3-7}
& & BiLSTM          & 128   & 2 & 0.73  & 676\\ 
& & BiLSTM + Attn   & 128   & 2 & {\color{blue}\bf 0.75}  & 801\\ 
\cmidrule{3-7}
& & GRU             & 128   & 2 & 0.71  & 600\\
& & GRU + Attn      & 128   & 2 & 0.73  & 591\\ 
\cmidrule{3-7}
& & BiGRU           & 128   & 2 & 0.73  & 721\\
& & BiGRU + Attn    & 128   & 2 & 0.74  & 740\\
\cmidrule{3-7}
& & BiGRU           & 1280  & 2 & 0.74  & 1980\\
& & BiGRU + Attn    & 1280  & 2 & {\color{blue}\bf 0.75} & 1300\\
\cmidrule{3-7}
& & BiGRU           & 128   & 4 & 0.71  & 1012\\
& & BiGRU + Attn    & 128   & 4 & {\color{blue}\bf 0.75}  & 1144\\ 
\midrule
\multirow{8}{*}{SciBERT} & \multirow{8}{*}{\begin{tabular}[c]{@{}c@{}}61440\\($768$)\end{tabular}}& LSTM & 128 & 2 & 0.70 & 619\\
& & LSTM + Attn     & 128   & 2 & 0.69  & 431\\
\cmidrule{3-7}
& & BiLSTM          & 128   & 2 & 0.72  & 949\\
& & BiLSTM + Attn   & 128   & 2 & 0.71  & 557\\
\cmidrule{3-7}
& & GRU             & 128   & 2 & 0.70  & 496\\
& & GRU + Attn      & 128   & 2 & 0.69  & 412\\
\cmidrule{3-7}
& & BiGRU           & 128   & 2 & 0.72  & 698\\
& & BiGRU + Attn    & 128   & 2 & 0.71  & 537\\
\midrule
\multirow{8}{*}{FastText} & \multirow{8}{*}{\begin{tabular}[c]{@{}c@{}}8000\\($100$)\end{tabular}} & 
    LSTM            & 128   & 2 & 0.65  & 560\\
& & LSTM  + Attn    & 128   & 2 & {\color{blue}\bf 0.75}  & 748\\
\cmidrule{3-7}
& & BiLSTM          & 128   & 2 & 0.68  & 658\\
& & BiLSTM + Attn   & 128   & 2 & {\color{blue}\bf 0.76}  & 862\\
\cmidrule{3-7}
& & GRU             & 128   & 2 & 0.69  & 496\\
& & GRU + Attn      & 128   & 2 & {0.74}  & 752\\
\cmidrule{3-7}
& & BiGRU           & 128   & 2 & 0.68  & 740\\
& & BiGRU + Attn    & 128   & 2 & {\color{blue}\bf 0.76}  & 852\\ 
\midrule
CCNN                                                                   & 1014                                                                                             & CNN                  & 1008  & 6      & 0.68     & 560               \\ 
\midrule
\multirow{2}{*}{\begin{tabular}[c]{@{}c@{}}USE \end{tabular}} & \multirow{2}{*}{512}                                                                             & \multirow{2}{*}{MLP} & 1024  & 4      & 0.63     & 729               \\
                                                                                  &                                                                                                  &                      & 2048  & 4      & 0.64     & 766               \\
\bottomrule
\end{tabular}
\end{table*}